\documentclass[a4paper,10pt]{article}
\usepackage[dvips]{graphicx}
\usepackage{theorem}
\usepackage{subeqn}
\usepackage{floatflt}
\newcommand{\gbf}[1] {\mbox{\boldmath${#1}$\unboldmath}}
\newcommand{\be}{\begin{equation}}
\newcommand{\ee}{\end{equation}}
\newcommand{\beq}{\begin{equation}}
\newcommand{\eeq}{\end{equation}}
\newcommand{\bed}{\begin{displaymath}}
\newcommand{\eed}{\end{displaymath}}
\newcommand{\beqa}{\begin{eqnarray}}
\newcommand{\eeqa}{\end{eqnarray}}
\newcommand{\beqann}{\begin{eqnarray*}}
\newcommand{\eeqann}{\end{eqnarray*}}
\newcommand{\bseq}{\begin{subequations}}
\newcommand{\eseq}{\end{subequations}}

\newcommand{\ba}{\begin{array}}
\newcommand{\ea}{\end{array}}

\newcommand{\set}[3]{\{\,#1\,\}_{#2}^{#3}}

\newcommand{\M}{{\bf M}}

\newcommand{\negr}[1]{{\bf {#1}}}
\newcommand{\nigr}[2]{{\bf {#1}}_{#2}}

 \makeatletter
\theoremstyle{plain}

\newtheorem{Def}{Definition}

\def\@normalsize{\@setsize\normalsize{10pt}\xpt\@xpt
\abovedisplayskip 3pt plus 1pt minus 3pt
\belowdisplayskip\abovedisplayskip \abovedisplayshortskip \z@ plus
3pt \belowdisplayshortskip 3pt plus 2pt minus 3pt
\let\@listi\@listI}
\makeatother
\begin{document}
\date{}  % Il n' a pas de date
%%%%%%%%%%%%%%%%%%%%%%%%%%%%%%%%%%%%%%%%%%%%%%%%
%  Titre
%%%%%%%%%%%%%%%%%%%%%%%%%%%%%%%%%%%%%%%%%%%%%%%%
\title {\noindent\bf The Computation of All 4R Serial Spherical Wrists
With an Isotropic Architecture}
%%%%%%%%%%%%%%%%%%%%%%%%%%%%%%%%%%%%%%%%%%%%%%%%
%for two authors (this is what is printed)
%%%%%%%%%%%%%%%%%%%%%%%%%%%%%%%%%%%%%%%%%%%%%%%%
\author{\begin{tabular}[t]{c}
 {Damien Chablat$^1$ and Jorge Angeles$^2$} \\
 {\em $^1$Institut de Recherche en Communications et Cybern\'etique de Nantes
 \footnote{IRCCyN: UMR n$^\circ$ 6597 CNRS, \'Ecole Centrale de Nantes,
                     Universit\'e de Nantes, \'Ecole des Mines de Nantes}} \\
 {\em 1, rue de la No\"e, 44321 Nantes, France} \\
 {\em $^2$Department of Mechanical Engineering $\&$} \\
 {\em Centre for Intelligent Machines, McGill University} \\
 {\em  817 Sherbrooke Street West, Montreal, Canada H3A 2K6} \\
 {\bf Damien.Chablat@irccyn.ec-nantes.fr, angeles@cim.mcgill.ca}
\end{tabular}}
%%%%%%%%%%%%%%%%%%%%%%%%%%%%%%%%%%%%%%%%%%%%%%%%
%Creation du titre de l'article
%%%%%%%%%%%%%%%%%%%%%%%%%%%%%%%%%%%%%%%%%%%%%%%%
\maketitle
%%\thispagestyle{empty}
%%%%%%%%%%%%%%%%%%%%%%%%%%%%%%%%%%%%%%%%%%%%%%%%
%   Abstract
%%%%%%%%%%%%%%%%%%%%%%%%%%%%%%%%%%%%%%%%%%%%%%%%
{\noindent\bf Abstract:} {A spherical wrist of the serial type is
said to be isotropic if it can attain a posture whereby the
singular values of its Jacobian matrix are all identical and
nonzero. What isotropy brings about is robustness to
manufacturing, assembly, and measurement errors, thereby
guaranteeing a maximum orientation accuracy. In this paper we
investigate the existence of redundant isotropic architectures,
which should add to the dexterity of the wrist under design by
virtue of its extra degree of freedom. The problem formulation
leads to a system of eight quadratic equations with eight
unknowns. The Bezout number of this system is thus $2^8 = 256$,
its BKK bound being $192$. However, the actual number of solutions
is shown to be $32$. We list all solutions of the foregoing
algebraic problem. All these solutions are real, but distinct
solutions do not necessarily lead to distinct manipulators. Upon
discarding those algebraic solutions that yield no new wrists, we
end up with exactly eight distinct architectures, the eight
corresponding manipulators being displayed at their isotropic
posture. }
%%%%%%%%%%%%%%%%%%%%%%%%%%%%%%%%%%%%%%%%%%%%%%%%
\section{Introduction}
%%%%%%%%%%%%%%%%%%%%%%%%%%%%%%%%%%%%%%%%%%%%%%%%
The kinematic design of redundant spherical wrists under isotropy
conditions is the subject of this paper. A manipulator is call
{\em isotropic} if its Jacobian matrix can attain isotropic values
on certain postures \cite{Salisbury:82}. A matrix, in turn, is
called {\em isotropic} if its singular values are all identical
and nonzero. Furthermore, the matrix condition number  can be
defined as the ratio of its greatest to its smallest singular
values \cite{Golub:89}. Thus, isotropic matrices have a minimum
condition number of unity. The kinematic structure of industrial
manipulators are frequently decoupled into a positioning and an
orientation submanipulator. The latter is designed with revolute
joints whose axes intersect. However, when these three joints are
coplanar, the manipulator becomes singular. As a means to cope
with singularities, redundant wrists have been suggested
\cite{Long:89}. An extensive bibliography on the design of
spherical wrists can be found in \cite{Farhang:99}.

Prior to our analysis leading to the architectures sought, we
recall a few geometric concepts in the section below.
%%%%%%%%%%%%%%%%%%%%%%%%%%%%%%%%%%%%%%%%%%%%%%%%
\section{Isotropic Sets of Points on the Unit Sphere}\label{s:isopts}
%%%%%%%%%%%%%%%%%%%%%%%%%%%%%%%%%%%%%%%%%%%%%%%%
Consider the set ${\cal S}\equiv \set{P_k}1n$ of $n$ points on the
unit sphere, of position vectors $\set{\negr e_k}1n$. Apparently,
all the vectors of the foregoing set are of unit Euclidean norm.
The {\em second-moment tensor} $\negr H$ of ${\cal S}$ is defined
as
 \beqa
   \negr H = \sum_1^n \negr e_k \negr e_k^T
 \eeqa
The set ${\cal S}$ is said to be {\em isotropic} if and only if
its second-moment tensor is isotropic. Since \negr H is symmetric
and positive-definite, it is isotropic if its matrix
representation is proportional to the $3\times3$ identity matrix
\negr 1, the proportionality factor, denoted here with $\sigma^2$,
being the square of the triple singular value of \negr H. In our
case, apparently, the singular values of \negr H coincide with its
eigenvalues.

We note that, if $\cal S$ is the set of vertices of a Platonic solid,
then \negr H is isotropic. Table~\ref{Table:Platonic-solids} records
the values of $n$ and $\sigma$ for each Platonic solid.
 \begin{table}[hbt]
  \begin{center}
  \begin{tabular}{|c|c|c|c|c|c|c|c|c|} \hline
    ~~           &
    Tetrahedron  &
    Cube         &
    Octahedron   &
    Icosahedron  &
    Dodecahedron \\ \hline
    $n$          &
    $4$          &
    $8$          &
    $6$          &
    $12$         &
    $20$         \\ \hline
    $\sigma$     &
    $4/3$        &
    $8/3$        &
    $2$          &
    $4$          &
    $20/3$       \\ \hline
  \end{tabular}
  \end{center}
  \caption{The values of $n$ and $\sigma$ for the Platonic solids}
  \label{Table:Platonic-solids}
 \end{table}

\noindent{\bf Remark~1:} It is apparent that, if a point $P_k$ of an
    arbitrary set $\cal S$ of points on the unit sphere is replaced by
    its {\em antipodal} $Q_k$, of position vector $\negr q_k = -\negr
    e_k$, then the second-moment tensor \negr H of $\cal S$ is {\em
    preserved}.

The replacement of a point on the unit sphere by its antipodal will be
termed, henceforth, {\em antipodal exchange}. As a consequence of
Remark~1, then, the isotropy of a set of points on the unit sphere is
preserved under any antipodal exchanges.

We started by recalling the second moment of a set of points on the
unit sphere because this is simpler to handle than the corresponding
first moment. Besides, in deriving isotropic spherical wrists, all we
need is the second moment. The first moment of a set of points on the
unit sphere is somewhat more elusive, because the {\em centroid} of
the set must be a point on the unit sphere as well. Thus, while the
second moment \negr H was taken with respect to the center of the
sphere, the first moment, when taken with respect to the centroid,
must vanish. The centroid of the set not being of interest to us in
the context of spherical-wrist design, it will be left out of the
discussion.
%%%%%%%%%%%%%%%%%%%%%%%%%%%%%%
\subsection{Trivial Isotropic Sets of Points on the Unit Sphere}
%%%%%%%%%%%%%%%%%%%%%%%%%%%%%%
The simplest sets of isotropic points are thus the sets of
vertices of the Platonic solids. Hence,
 \begin{Def}[Trivial isotropic set]
An isotropic set ${\cal S}$ of $n$ points on the unit sphere is
called {\em trivial} if it consists of the set of vertices of a
Platonic solid (inscribed, of course, on the unit sphere.)
 \end{Def}
It is noteworthy that isotropic sets on the unit sphere exist that
are none of the Platonic solids, e.g., the 64-vertex polyhedron
defined by the molecule of the buckminsterfullerene, popularly
known as the Buckyball. The name comes from the architect R.
Buckminster Fuller, who used this polyhedron as the structure of
the geodesic dome built on occasion of the Universal Exhibit of
1967 in Montreal. This polyhedron is also present in the patterns
of soccer balls.

Also note that the set of points on the unit sphere leading to an
isotropic architecture for a spherical wrist need not be laid out
with the center of the sphere as its centroid, which is a
condition found for points in the plane \cite{Angeles:00}. For
example, the three points of intersection of the unit sphere with
the axes of an orthogonal coordinate frame with origin at the
center of the sphere define an isotropic spherical wrist, namely,
the one most commonly encountered in commercial manipulators, yet
the above set of points corresponds to none of the Platonic
solids. This three-revolute wrist is termed {\em orthogonal}
because its neighboring axes make right angles.

Trivial sets of isotropic points are important because they allow the
derivation of nontrivial sets by simple operations, as described
below.
%%%%%%%%%%%%%%%%%%%%%%%%%%%%%%%%%%%%%%%%%%%%%%%%
\subsection{Properties of Isotropic Sets of Points on the Unit Sphere}
%%%%%%%%%%%%%%%%%%%%%%%%%%%%%%%%%%%%%%%%%%%%%%%%
First, note

\noindent{\bf Lemma~1:} An isotropic set ${\cal S}$ of points on
the unit sphere remains isotropic under any {\em isometric
transformation} of the set.

An isometry being either a rigid-body rotation or a reflection, the
foregoing lemma should be obvious. Moreover, rigid-body rotations of
isotropic sets are uninteresting because they amount to looking at the
given set from a different viewpoint. However, distinct isotropic sets
can be derived from reflections of isotropic sets about planes or
lines. Nevertheless, as shown in the Appendix, a reflection about a
line amounts to a rigid-body rotation about the line through an angle
of $\pi$. As a consequence, then, only reflections about planes will
be considered when defining new isotropic sets from trivial ones.
%%%%%%%%%%%%%%%%%%%%%%%%%%%%%%
\subsection{Nontrivial Isotropic Sets of Points}
%%%%%%%%%%%%%%%%%%%%%%%%%%%%%%
We show in this subsection, with a numerical example, how nontrivial
sets of isotropic points on the unit sphere can be derived from a
trivial set by application of reflections about planes. Now, since the
reflection plane can be defined in infinitely-many ways, a
correspondingly infinite number of reflections is possible. We are
interested only in {\em linearly-independent} reflections, which are,
apparently, only three, one about each of three mutually orthogonal
planes.

A trivially isotropic set of four points, namely, the vertices of a
regular tetrahedron inscribed in the unit sphere, is given below:
 \bseq
 \beq
   \negr e_1 =
   \left[\begin{array}{c}
           1 \\  0 \\ 0
         \end{array}
   \right],\quad
   \negr e_2 =
   \left[\begin{array}{c}
           -1/3 \\  -2 \sqrt{2} / 3 \\  0
         \end{array}
   \right],\quad
   \negr e_3 =
   \left[\begin{array}{c}
           -1/3 \\    \sqrt{2} / 3 \\  \sqrt{6}/3
         \end{array}
   \right],\quad
   \negr e_4 =
   \left[\begin{array}{c}
           -1/3 \\ \sqrt{2} / 3 \\  -\sqrt{6}/3
         \end{array}
   \right]
   \label{equation:trivial_set}
 \eeq
We produce now three nontrivial sets of isotropic points by
reflecting the foregoing set onto the three coordinate planes,
$y$-$z$, $x$-$z$ and $x$-$y$, successively. We display below the
three new sets:
\begin{enumerate}
\item The reflection with respect to the $y$-$z$ plane gives
 \beqa
   \negr e^\prime_1 =
   \left[\begin{array}{c}
          -1 \\  0 \\  0
         \end{array}
   \right],\quad
   \negr e^\prime_2 =
   \left[\begin{array}{c}
           1/3 \\ -2 \sqrt{2} / 3 \\  0
         \end{array}
   \right],\quad
   \negr e^\prime_3 =
   \left[\begin{array}{c}
           1/3 \\   \sqrt{2} / 3 \\ \sqrt{6}/3
         \end{array}
   \right],\quad
   \negr e^\prime_4 =
   \left[\begin{array}{c}
           1/3 \\ \sqrt{2} / 3 \\ -\sqrt{6}/3
         \end{array}
   \right]
 \eeqa
\item The reflection with respect to the $x$-$z$ plane gives
 \beqa
   \negr e^{\prime\prime}_1 =
   \left[\begin{array}{c}
           1 \\  0 \\  0
         \end{array}
   \right],\quad
   \negr e^{\prime\prime}_2 =
   \left[\begin{array}{c}
           -1/3 \\ 2 \sqrt{2} / 3 \\  0
         \end{array}
   \right],\quad
   \negr e^{\prime\prime}_3 =
   \left[\begin{array}{c}
           -1/3 \\   -\sqrt{2} / 3 \\ \sqrt{6}/3
         \end{array}
   \right],\quad
   \negr e^{\prime\prime}_4 =
   \left[\begin{array}{c}
           -1/3 \\ -\sqrt{2} / 3 \\ -\sqrt{6}/3
         \end{array}
   \right]
 \eeqa
\item The reflection  with respect to the $x$-$y$ plane gives
 \beqa
   \negr e^{\prime\prime\prime}_1 =
   \left[\begin{array}{c}
           1 \\  0 \\  0
         \end{array}
   \right],\quad
   \negr e^{\prime\prime\prime}_2 =
   \left[\begin{array}{c}
           -1/3 \\ -2 \sqrt{2} / 3 \\  0
         \end{array}
   \right],\quad
   \negr e^{\prime\prime\prime}_3 =
   \left[\begin{array}{c}
           -1/3 \\ \sqrt{2} / 3 \\ -\sqrt{6}/3
         \end{array}
   \right],\quad
   \negr e^{\prime\prime\prime}_4 =
   \left[\begin{array}{c}
           -1/3 \\ \sqrt{2} / 3 \\  \sqrt{6}/3
         \end{array}
   \right]
 \eeqa
 \end{enumerate}
 \eseq
%%%%%%%%%%%%%%%%%%%%%%%%%%%%%%%%%%%%%%%%%%%%%%%%
\section{Isotropic Spherical Wrists}
%%%%%%%%%%%%%%%%%%%%%%%%%%%%%%%%%%%%%%%%%%%%%%%%
Most serial wrists encountered in manipulators are provided with three
revolute joints. We start with a general $n$-revolute spherical wrist,
as depicted in Fig.~\ref{figure:n-dof-spherical-wrist}, with Jacobian
matrix \negr J given by \cite{Angeles:97}
% \begin{figure}[hbt]
\begin{floatingfigure}[r]{43mm}
\begin{center}
   \includegraphics[width=43mm,height=54mm]{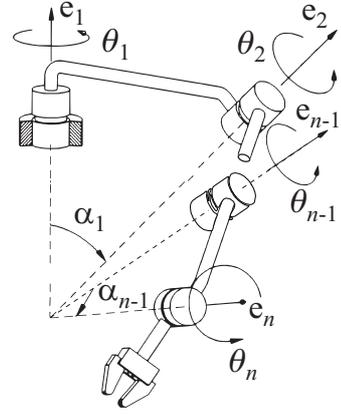}
   \caption{A general $n$-revolute spherical wrist}
   \protect\label{figure:n-dof-spherical-wrist}
\end{center}
\end{floatingfigure}
% \end{figure}
 \beqa \negr J =
  \left[\begin{array}{cccc}
   \negr e_1  & \negr e_2 & \cdots & \negr e_n
  \end{array}
  \right]
  \label{equation:jacobian_matrix}
 \eeqa
\noindent where, we recall, $\nigr ek$ is the unit vector
indicating the direction of the $k$th revolute axis. We display
below the kinematic relation between the joint-rate vector
$\dot{\gbf\theta}$ and the angular-velocity vector $\gbf\omega$ of
the end-effector (EE):
 \beq
  \negr J\dot{\gbf\theta} = {\gbf\omega}
  \label{e:kine}
 \eeq
It should be apparent that

\noindent{\bf Remark~2:} A set $\set{\nigr ek}1n$ of unit vectors
produces $n!$ Jacobian matrices, and hence, $n!$ {\em distinct
wrists}.

Kinetostatic isotropy requires that the singular values of the
Jacobian matrix be all identical and nonzero, i.e.,
 \beqa
    \negr J \negr J^T = \sigma^2 \negr 1
 \eeqa
where $\sigma$ is the common singular value, of multiplicity
three, and \negr 1 is, as defined earlier, the $3\times 3$
identity matrix. The isotropy condition thus leads to
 \beqa
   \sum^n_1 \negr e_k \negr e_k^T = \sigma^2 \negr 1
   \label{equation:sigma-e_i}
  \eeqa
The value of $\sigma$ is found by taking the trace of both sides of
eq.(\ref{equation:sigma-e_i}), which yields
 \beqa
   \sum^n_1 \negr e_k \cdot \negr e_k = 3 \sigma^2
 \eeqa
and hence,
 \beqa
   \sigma = \sqrt{\frac{n}{3}}\label{e:sigma}
 \eeqa
i.e., if $\negr J$ is isotropic, then (a) every pair of
 $n$-dimensional rows of \negr J is orthogonal and (b) the three rows
 of \negr J have the same Euclidean norm, namely, $\sqrt{n/3}$.

Now, since it does not appear practical to design wrists with more
than four revolutes, we limit ourselves, in the balance of the paper,
to four-revolute manipulators, i.e., we set $n=4$, unless otherwise
stated.

%%%%%%%%%%%%%%%%%%%%%%%%%%%%%%%%%%%%%%%%%%%%%%%%
\section{Four-Axis Isotropic Spherical Wrists}
%%%%%%%%%%%%%%%%%%%%%%%%%%%%%%%%%%%%%%%%%%%%%%%%
In this section we obtain all possible four-revolute serial
spherical wrists with isotropic architectures.
\par
%%%%%%%%%%%%%%%%%%%%%%%%%%%%%%%%%%%%%%%%%%%%%%%%
%\subsubsection{Isotropy conditions in algebraic form}
%%%%%%%%%%%%%%%%%%%%%%%%%%%%%%%%%%%%%%%%%%%%%%%%
The algebraic problem at hand consists in finding the set of
vectors $\set{\nigr ek}14$ that verify the isotropy conditions of
eq.(\ref{equation:sigma-e_i}). Without loss of generality, we
define $\nigr e1$ parallel to the $x$ axis of the coordinate frame
at hand; then, we let $\nigr e2$ lie in the $y$-$z$ plane of the
same frame, while the remaining two vectors are left arbitrary. We
thus have
 \beqa
   \negr e_1 =
   \left[\begin{array}{c}
           1 \\  0 \\  0
         \end{array}
   \right],\quad
   \negr e_2 =
   \left[\begin{array}{c}
           c \\  s \\  0
         \end{array}
   \right],\quad
   \negr e_3 =
   \left[\begin{array}{c}
           x \\ y \\ z
         \end{array}
   \right],\quad
   \negr e_4 =
   \left[\begin{array}{c}
           u \\ v \\  w
         \end{array}
   \right]
 \eeqa
in which $c\equiv\cos\alpha_1$ and $s\equiv\sin\alpha_1$, as per
 Fig.~\ref{figure:n-dof-spherical-wrist}. The isotropy condition
 (\ref{equation:sigma-e_i}), in terms of the foregoing components and
 with $\sigma^2 = 4/3$, yields, then,
\bseq
 \beqa
    1 + c^2 + x^2 + u^2 &=& 4/3\label{e:1+} \\
    s^2 + y^2 + v^2 &=& 4/3 \\
    z^2 + w^2 &=& 4/3 \\
    c s + x y + u v &=& 0\label{e:cs} \\
    z y + w v &=& 0 \\
    x z + u w &=& 0
  \eeqa
Besides, we have the normality of $\nigr e2$ and $\nigr e3$:
  \beqa
    c^2 + s^2 &=& 1\label{e:c&s} \\
    x^2 + y^2 + z^2 &=& 1\label{e:e_3}
  \eeqa
  \label{equation:systeme-1}
 \eseq
$\!\!$the normality of $\nigr e4$ being embedded in the foregoing
system of equations. Indeed, this is obtained upon adding
eqs.(\ref{equation:systeme-1}a--c) and subtracting this sum from
the sum of eqs.(\ref{e:c&s} \& h). We have now eight quadratic
equations for eight unknowns. The Bezout number of this system is
thus $2^8 = 256$, which means that up to 256 solutions are to be
expected, including real and complex, as well as multiple
solutions. Moreover, the BKK bound \cite{Emiris:94} of the same
system turns out to be $192$. It will be shown presently that this
number is too big, the total number of solutions being
substantially smaller. In order to find the solutions of interest,
we eliminate successively all the unknowns but $u$ to obtain a
monovariate polynomial in this unknown. First, we solve for $x$,
$y$ and $c$ from eqs.(\ref{e:cs}--f), thus obtaining
 \bseq
  \beqa
     x &=& -\frac{w u}{z}\label{e:x} \\
     y &=& -\frac{v w}{z}\label{e:y} \\
     c& =& \frac{u (w y - v z)}{s z}\label{e:c}
  \eeqa
 \eseq
Upon substituting eqs.(\ref{e:x}--c) into eqs.(\ref{e:1+}--c) and
 (\ref{e:c&s} \& b), we obtain
 \bseq
 \beqa
     w^2 u^2 + v^2 w^2 + z^4 - z^2 &=& 0\label{e:w^2}\\
    3\,{s}^{2}{z}^{2}+3\,{v}^{2}{w}^{2}+3\,{v}^{2}{z}^{2}-4\,{z}^{2}&=&0
    \label{e:3s^2z^2}\\
   3\,{u}^{2}{w}^{4}{v}^{2}+6\,{u}^{2}{w}^{2}{v}^{2}{z}^{2}+3\,{u}^{2}{v}
^{2}{z}^{4}+3\,{w}^{2}{u}^{2}{s}^{2}{z}^{2}+3\,{u}^{2}{z}^{4}{s}^{2}-{
s}^{2}{z}^{4} &=& 0\label{e:3u^2w^4v^2}\\
   {u}^{2}{w}^{4}{v}^{2}+2\,{u}^{2}{w}^{2}{v}^{2}{z}^{2}+{u}^{2}{v}^{2}{z
}^{4}+{s}^{4}{z}^{4}-{s}^{2}{z}^{4}  &=& 0\label{e:u^2w^4v^2}\\
   3\,{z}^{2}+3\,{w}^{2}-4  &=& 0\label{e:3z^2}
   \eeqa
   \label{equation:systeme-2}
 \eseq
It is noteworthy that the system of equations
(\ref{equation:systeme-2}a--e) contains only second and fourth
powers of all the unknowns, which allows for a recursive solution,
as we shall show below. First, from eq.(\ref{equation:systeme-2}e)
we solve for $w$:
 \bseq
   \beq
      w= \pm \frac{1}{3} \sqrt {12-9\,{z}^{2}}~~~~
      \label{equation:solution-w}
   \eeq
Upon substitution of eq.(\ref{equation:solution-w}) into
eq.(\ref{equation:systeme-2}b) we obtain $s$:
   \beq
      s= \pm \frac{2}{3} {\frac {\sqrt {3\,{z}^{2}-3\,{v}^{2}}}{z}}
      \label{equation:solution-s}
   \eeq
Likewise, substitution of eq.(\ref{equation:solution-s}) into
eq.(\ref{equation:systeme-2}a) yields $v$:
   \beq
      v= \pm \sqrt {{z}^{2}-4\,{u}^{2}}~~~~~
      \label{equation:solution-v}
   \eeq
Finally, substitution of eq.(\ref{equation:solution-v}) into
eq.(\ref{equation:systeme-2}d) leads to $z$:
   \beqa
     z= \pm \sqrt {6}u\label{e:sol-z}
   \eeqa
Now, substitution of eqs.(\ref{equation:solution-w}-d) into
eq.(\ref{e:3u^2w^4v^2}) leads to a monovariate polynomial:
   \beqa
     u (3\,u-1) (3\,u + 1) = 0\label{e:u_3}
   \eeqa
 \eseq
The above equation reduces, in fact, to a quadratic equation
because $u$ cannot vanish, as we shall show presently. Thus, the
two possible solutions for eq.(\ref{e:u_3}) are
 \beq
  u=\pm \frac{1}{3}\label{e:u}
 \eeq
We thus have a set of five quadratic expressions for the five
unknowns $u$, $z$, $v$, $s$ and $w$, which means that we have
found $2^5=32$ distinct solutions, as displayed in
Table~\ref{Table:Solutions}. Therefore, the Bezout number of this
system overestimates the number of solutions by a factor of eight,
while the BKK bound by a factor of six.

\begin{table}[!hb]
  \begin{center}
  \begin{tabular}{|r|r|r|r|r|r|r|r|r|r|} \hline
Sol'n \#& $c$~ & $s$~~~ & $x$~~ & $y$~~ & $z$~~ & $u$~ & $v$~~ &
$w$~~
\\ \hline
1&   $ 1/3$ & $-2\sqrt{2}/3$ & $-1/3$ & $-\sqrt{2}/3$ & $
\sqrt{6}/3$ & $ 1/3$ & $ \sqrt{2}/3$ & $ \sqrt{6}/3$ \\ \hline 2&
$ 1/3$ & $-2\sqrt{2}/3$ & $-1/3$ & $-\sqrt{2}/3$ & $ \sqrt{6}/3$ &
$-1/3$ & $-\sqrt{2}/3$ & $-\sqrt{6}/3$ \\ \hline 3&   $ 1/3$ &
$-2\sqrt{2}/3$ & $-1/3$ & $-\sqrt{2}/3$ & $-\sqrt{6}/3$ & $ 1/3$ &
$ \sqrt{2}/3$ & $-\sqrt{6}/3$ \\ \hline 4&   $ 1/3$ &
$-2\sqrt{2}/3$ & $-1/3$ & $-\sqrt{2}/3$ & $-\sqrt{6}/3$ & $-1/3$ &
$-\sqrt{2}/3$ & $ \sqrt{6}/3$ \\ \hline 5&   $ 1/3$ &
$-2\sqrt{2}/3$ & $ 1/3$ & $ \sqrt{2}/3$ & $ \sqrt{6}/3$ & $ 1/3$ &
$ \sqrt{2}/3$ & $-\sqrt{6}/3$ \\ \hline 6&   $ 1/3$ &
$-2\sqrt{2}/3$ & $ 1/3$ & $ \sqrt{2}/3$ & $ \sqrt{6}/3$ & $-1/3$ &
$-\sqrt{2}/3$ & $ \sqrt{6}/3$ \\ \hline 7&   $ 1/3$ &
$-2\sqrt{2}/3$ & $ 1/3$ & $ \sqrt{2}/3$ & $-\sqrt{6}/3$ & $ 1/3$ &
$ \sqrt{2}/3$ & $ \sqrt{6}/3$ \\ \hline 8&   $ 1/3$ &
$-2\sqrt{2}/3$ & $ 1/3$ & $ \sqrt{2}/3$ & $-\sqrt{6}/3$ & $-1/3$ &
$-\sqrt{2}/3$ & $-\sqrt{6}/3$ \\ \hline

9&   $ 1/3$ & $ 2\sqrt{2}/3$ & $-1/3$ & $ \sqrt{2}/3$ & $
\sqrt{6}/3$ & $ 1/3$ & $-\sqrt{2}/3$ & $ \sqrt{6}/3$ \\ \hline 10&
$ 1/3$ & $ 2\sqrt{2}/3$ & $-1/3$ & $ \sqrt{2}/3$ & $ \sqrt{6}/3$ &
$-1/3$ & $ \sqrt{2}/3$ & $-\sqrt{6}/3$ \\ \hline 11&  $ 1/3$ & $
2\sqrt{2}/3$ & $-1/3$ & $ \sqrt{2}/3$ & $-\sqrt{6}/3$ & $ 1/3$ &
$-\sqrt{2}/3$ & $-\sqrt{6}/3$ \\ \hline 12&  $ 1/3$ & $
2\sqrt{2}/3$ & $-1/3$ & $ \sqrt{2}/3$ & $-\sqrt{6}/3$ & $-1/3$ & $
\sqrt{2}/3$ & $ \sqrt{6}/3$ \\ \hline 13&  $ 1/3$ & $ 2\sqrt{2}/3$
& $ 1/3$ & $-\sqrt{2}/3$ & $ \sqrt{6}/3$ & $ 1/3$ & $-\sqrt{2}/3$
& $-\sqrt{6}/3$ \\ \hline 14&  $ 1/3$ & $ 2\sqrt{2}/3$ & $ 1/3$ &
$-\sqrt{2}/3$ & $ \sqrt{6}/3$ & $-1/3$ & $ \sqrt{2}/3$ & $
\sqrt{6}/3$ \\ \hline 15&  $ 1/3$ & $ 2\sqrt{2}/3$ & $ 1/3$ &
$-\sqrt{2}/3$ & $-\sqrt{6}/3$ & $ 1/3$ & $-\sqrt{2}/3$ & $
\sqrt{6}/3$ \\ \hline 16&  $ 1/3$ & $ 2\sqrt{2}/3$ & $ 1/3$ &
$-\sqrt{2}/3$ & $-\sqrt{6}/3$ & $-1/3$ & $ \sqrt{2}/3$ &
$-\sqrt{6}/3$ \\ \hline

17&  $-1/3$ & $-2\sqrt{2}/3$ & $-1/3$ & $ \sqrt{2}/3$ & $
\sqrt{6}/3$ & $ 1/3$ & $-\sqrt{2}/3$ & $ \sqrt{6}/3$ \\ \hline 18&
$-1/3$ & $-2\sqrt{2}/3$ & $-1/3$ & $ \sqrt{2}/3$ & $ \sqrt{6}/3$ &
$-1/3$ & $ \sqrt{2}/3$ & $-\sqrt{6}/3$ \\ \hline 19&  $-1/3$ &
$-2\sqrt{2}/3$ & $-1/3$ & $ \sqrt{2}/3$ & $-\sqrt{6}/3$ & $-1/3$ &
$ \sqrt{2}/3$ & $ \sqrt{6}/3$ \\ \hline 20&  $-1/3$ &
$-2\sqrt{2}/3$ & $-1/3$ & $ \sqrt{2}/3$ & $-\sqrt{6}/3$ & $ 1/3$ &
$-\sqrt{2}/3$ & $-\sqrt{6}/3$ \\ \hline 21&  $-1/3$ &
$-2\sqrt{2}/3$ & $ 1/3$ & $-\sqrt{2}/3$ & $ \sqrt{6}/3$ & $ 1/3$ &
$-\sqrt{2}/3$ & $-\sqrt{6}/3$ \\ \hline 22&  $-1/3$ &
$-2\sqrt{2}/3$ & $ 1/3$ & $-\sqrt{2}/3$ & $ \sqrt{6}/3$ & $-1/3$ &
$ \sqrt{2}/3$ & $ \sqrt{6}/3$ \\ \hline 23&  $-1/3$ &
$-2\sqrt{2}/3$ & $ 1/3$ & $-\sqrt{2}/3$ & $-\sqrt{6}/3$ & $-1/3$ &
$ \sqrt{2}/3$ & $-\sqrt{6}/3$ \\ \hline 24&  $-1/3$ &
$-2\sqrt{2}/3$ & $ 1/3$ & $-\sqrt{2}/3$ & $-\sqrt{6}/3$ & $ 1/3$ &
$-\sqrt{2}/3$ & $ \sqrt{6}/3$ \\ \hline

25&  $-1/3$ & $ 2\sqrt{2}/3$ & $-1/3$ & $-\sqrt{2}/3$ &
$-\sqrt{6}/3$ & $ 1/3$ & $ \sqrt{2}/3$ & $-\sqrt{6}/3$ \\ \hline
26&  $-1/3$ & $ 2\sqrt{2}/3$ & $-1/3$ & $-\sqrt{2}/3$ &
$-\sqrt{6}/3$ & $-1/3$ & $-\sqrt{2}/3$ & $ \sqrt{6}/3$ \\ \hline
27&  $-1/3$ & $ 2\sqrt{2}/3$ & $-1/3$ & $-\sqrt{2}/3$ & $
\sqrt{6}/3$ & $-1/3$ & $-\sqrt{2}/3$ & $-\sqrt{6}/3$ \\ \hline 28&
$-1/3$ & $ 2\sqrt{2}/3$ & $-1/3$ & $-\sqrt{2}/3$ & $ \sqrt{6}/3$ &
$ 1/3$ & $ \sqrt{2}/3$ & $ \sqrt{6}/3$ \\ \hline 29&  $-1/3$ & $
2\sqrt{2}/3$ & $ 1/3$ & $ \sqrt{2}/3$ & $-\sqrt{6}/3$ & $-1/3$ &
$-\sqrt{2}/3$ & $-\sqrt{6}/3$ \\ \hline 30&  $-1/3$ & $
2\sqrt{2}/3$ & $ 1/3$ & $ \sqrt{2}/3$ & $-\sqrt{6}/3$ & $ 1/3$ & $
\sqrt{2}/3$ & $ \sqrt{6}/3$ \\ \hline 31&  $-1/3$ & $ 2\sqrt{2}/3$
& $ 1/3$ & $ \sqrt{2}/3$ & $ \sqrt{6}/3$ & $-1/3$ & $-\sqrt{2}/3$
& $ \sqrt{6}/3$ \\ \hline 32& $-1/3$ & $ 2\sqrt{2}/3$ & $ 1/3$ & $
\sqrt{2}/3$ & $ \sqrt{6}/3$ & $ 1/3$ & $ \sqrt{2}/3$ &
$-\sqrt{6}/3$ \\ \hline
  \end{tabular}
  \end{center}
  \caption{The $32$ isotropic sets of four points on the unit sphere}
  \label{Table:Solutions}
 \end{table}
\par
The solutions can now be readily computed recursively. Indeed,
eqs.(\ref{equation:solution-w}--c) lead to
 \bseq
 \beqa
   v&=&\pm\sqrt{2}\, u\label{e:v&u}\\
   s&=&\pm\frac{2\sqrt{2}}{3}\label{e:s}\\
   w&=&\pm\frac{1}{3}\sqrt{6(2 - 9 u^2)}\label{e:w&u}
 \eeqa
 \eseq
the remaining unknowns being computed from eqs.(\ref{e:sol-z}),
(\ref{e:x}), (\ref{e:y}) and (\ref{e:c}), in this order.
\par
Now we show that none of the unknowns can vanish. We do this by
noting that:
\begin{enumerate}
\item If $u=0$, then $v=0$ by virtue of eq.(\ref{e:v&u}), while
  $w=\pm2\sqrt{3}/3$, by virtue of eq.(\ref{e:w&u}), thereby violating
  the normality condition on $\nigr e4$. Also note that $u=0$ leads to
  $c=0$ by virtue of eq.(\ref{e:c}), but this value, along with that
  of $s$ given by eq.(\ref{e:s}), violates the normality condition
  (\ref{e:c&s}); \label{it:u=0}
\item If $v=0$, then $u=0$ by virtue of eq.(\ref{e:v&u}), but,
  according to item~\ref{it:u=0}, this is impossible;\label{it:v=0}
\item If $w=0$, then, by virtue of eqs.(\ref{e:x} \& b), $x=y=0$;
  additionally, by virtue of eqs.(\ref{e:sol-z}) and (\ref{e:u}), $z=\pm
  \sqrt{6}/3$, thereby violating the normality condition
  (\ref{e:e_3});\label{it:w=0}
\item If $x=0$, then, according to eq.(\ref{e:sol-z}), either $u=0$ or
  $w=0$, but none of these can vanish, according to items~\ref{it:u=0}
  and \ref{it:w=0};\label{it:x=0}
\item If $y=0$, then, by virtue of eq.(\ref{e:y}) either $v=0$ or
  $w=0$, but these alternatives are not plausible according to
  items~\ref{it:v=0} and \ref{it:w=0};\label{it:y=0}
\item If $z=0$, then $u=0$, according with eq.(\ref{e:sol-z}), but, by
  virtue of item~\ref{it:u=0}, this is not possible;\label{it:z=0}
\item If $c=0$, then the normality of $\negr e_2$ requires that $s=\pm 1$,
  but this is impossible by virtue of eq.(~\ref{e:s});\label{it:c=0}
\item From eq.(\ref{e:s}), $s$ cannot vanish, thereby showing that
  none of the unknowns can vanish.
\end{enumerate}

Note that the trivial isotropic set of points given in
eq.(\ref{equation:trivial_set}) is solution 18 of the
Table~\ref{Table:Solutions}. Table~\ref{Table:Antipodal} records
seven isotropic sets of points obtained by means of antipodal
exchanges.

\begin{table}[!ht]
  \begin{center}
     \begin{tabular}{|c|c|c|c|c|c|c|c|c|c|c|c|c|} \hline
       $~$         & $P_2$ & $P_3$ & $P_4$ & $P_2P_3$ & $P_2P_4$ & $P_2P_4$ & $P_2P_3P_4$ \\ \hline
       Solution \# &   10  &   23  &   17  &   16    &    24    &     9    &     15      \\ \hline
     \end{tabular}
  \caption{Isotropic sets of points obtained with antipodal exchanges}
  \label{Table:Antipodal}
  \end{center}
\end{table}

With the forgoing eight isotropic sets of points, we obtain
additional isotropic sets by reflections onto the coordinate
planes $x$-$z$, $x$-$y$, $x$-$z$ and $x$-$y$; we can then verify
that these solutions are listed in Table~\ref{Table:Solutions}.
The corresponding solutions are given in
Table~\ref{Table:Reflexion}.
\begin{table}[!ht]
  \begin{center}
     \begin{tabular}{|c|c|c|c|c|c|c|c|c|c|c|c|c|} \hline
       Solution \#                             & 18 & 10 & 23 & 17 & 16 & 24 & 9  & 15 \\ \hline
       Reflection plane: $x$-$y$               & 19 & 12 & 22 & 20 & 14 & 21 & 11 & 13 \\ \hline
       Reflection plane: $x$-$z$               & 27 & 2  & 29 & 28 & 8  & 30 & 1  & 7  \\ \hline
       Reflection planes: $x$-$z$ and $x$-$y$  & 26 & 4  & 31 & 25 & 6  & 32 & 3  & 5  \\ \hline
     \end{tabular}
  \caption{Isotropic sets of points defined by reflections onto the coordinate planes $x$-$z$ and $x$-$y$}
  \label{Table:Reflexion}
  \end{center}
\end{table}

It is noteworthy that Table~4 includes a reflection about the
$x$-$y$ and the $x$-$z$ planes, which amount to a 180$^{\circ}$
rotation about the $x$ axis, and does not lead to a new wrist.

\begin{figure}[!hb]
    \begin{center}
    \begin{tabular}{cccc}
       \begin{minipage}[t]{33 mm}
           \centerline{\hbox{\includegraphics[width=37mm,height=41mm]{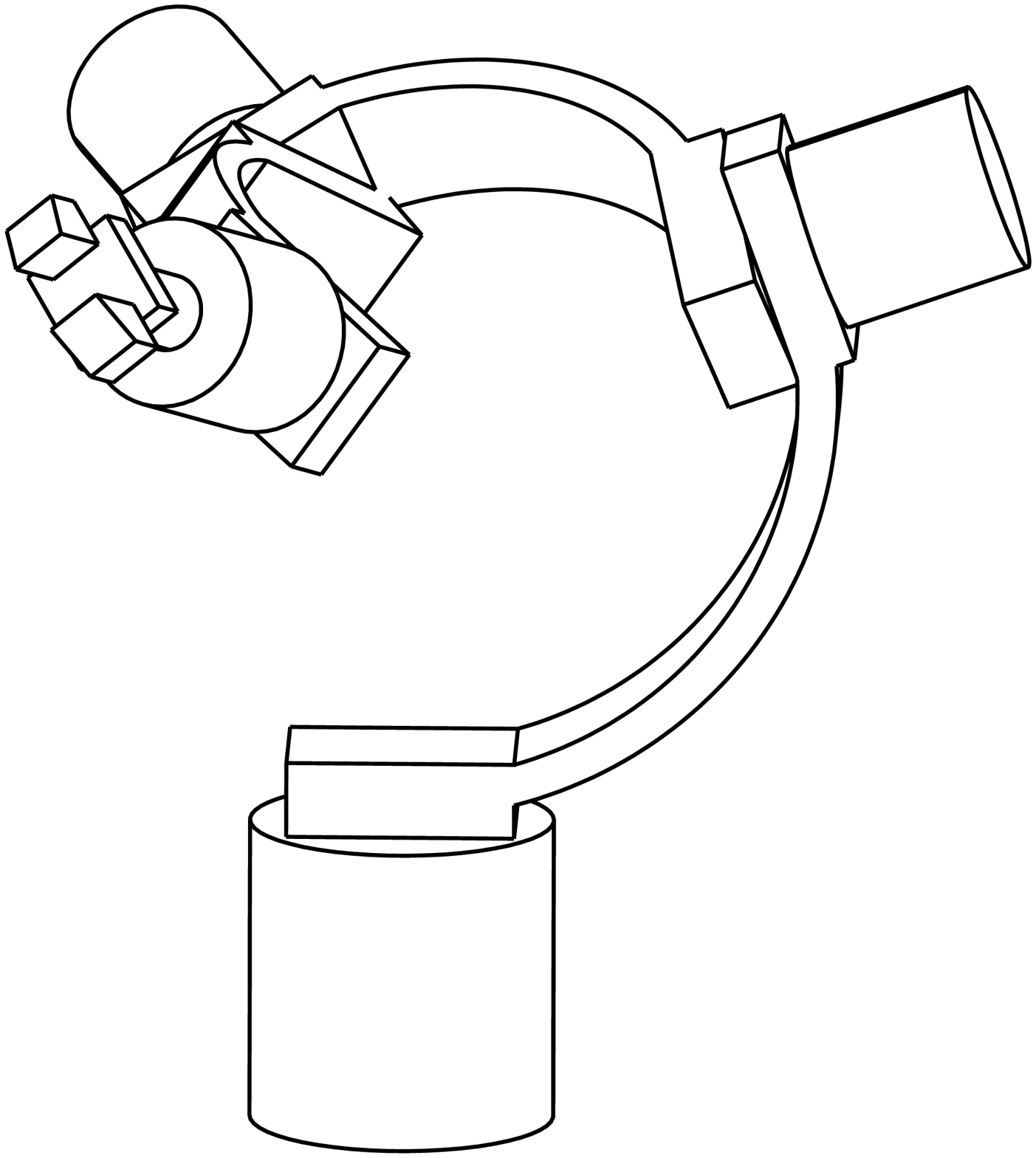}}}
           \centerline{\hbox{(a)}}
       \end{minipage} &
       \begin{minipage}[t]{33 mm}
           \centerline{\hbox{\includegraphics[width=37mm,height=32mm]{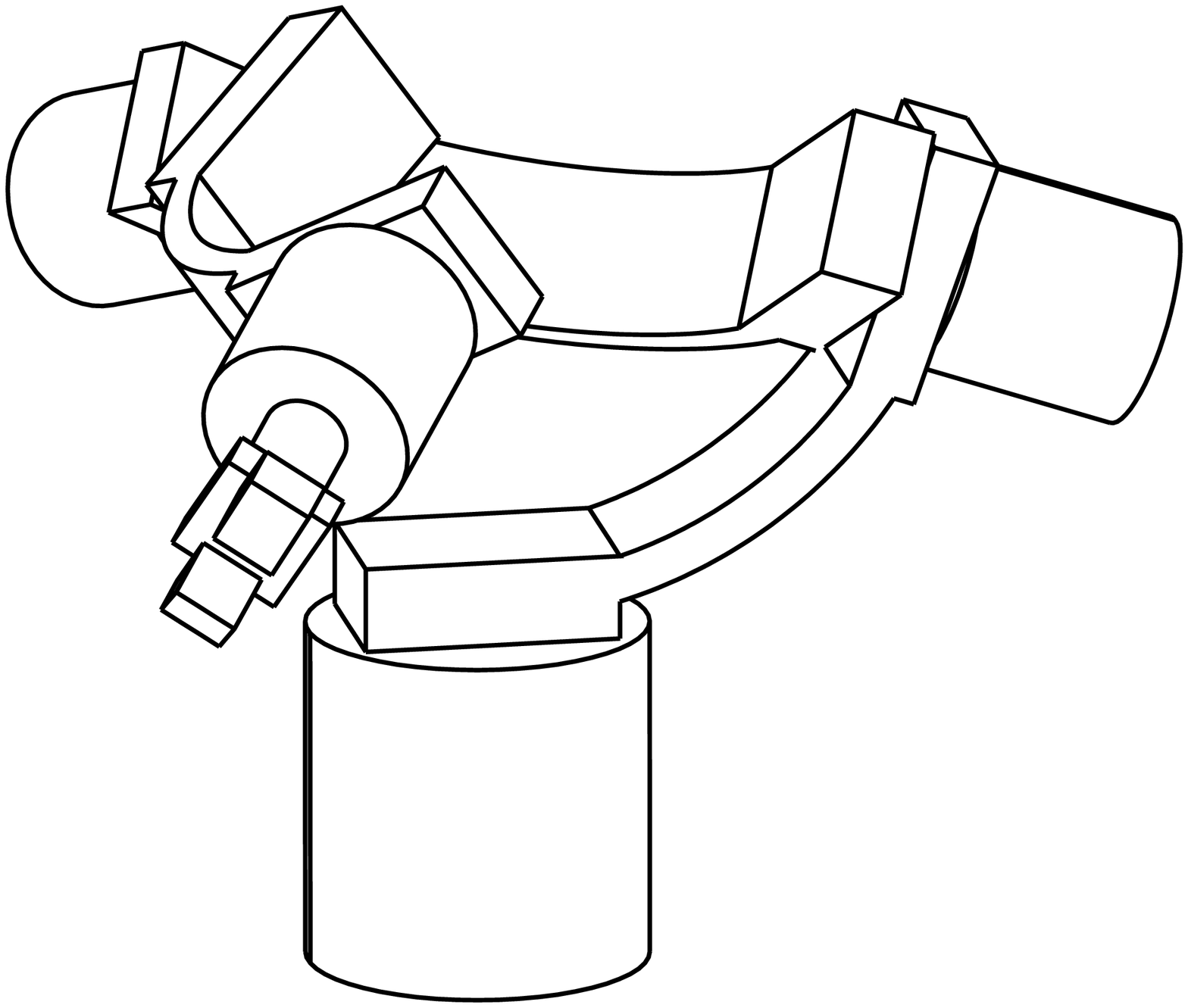}}}
           \centerline{\hbox{(b)}}
       \end{minipage} &
       \begin{minipage}[t]{33 mm}
           \centerline{\hbox{\includegraphics[width=36mm,height=42mm]{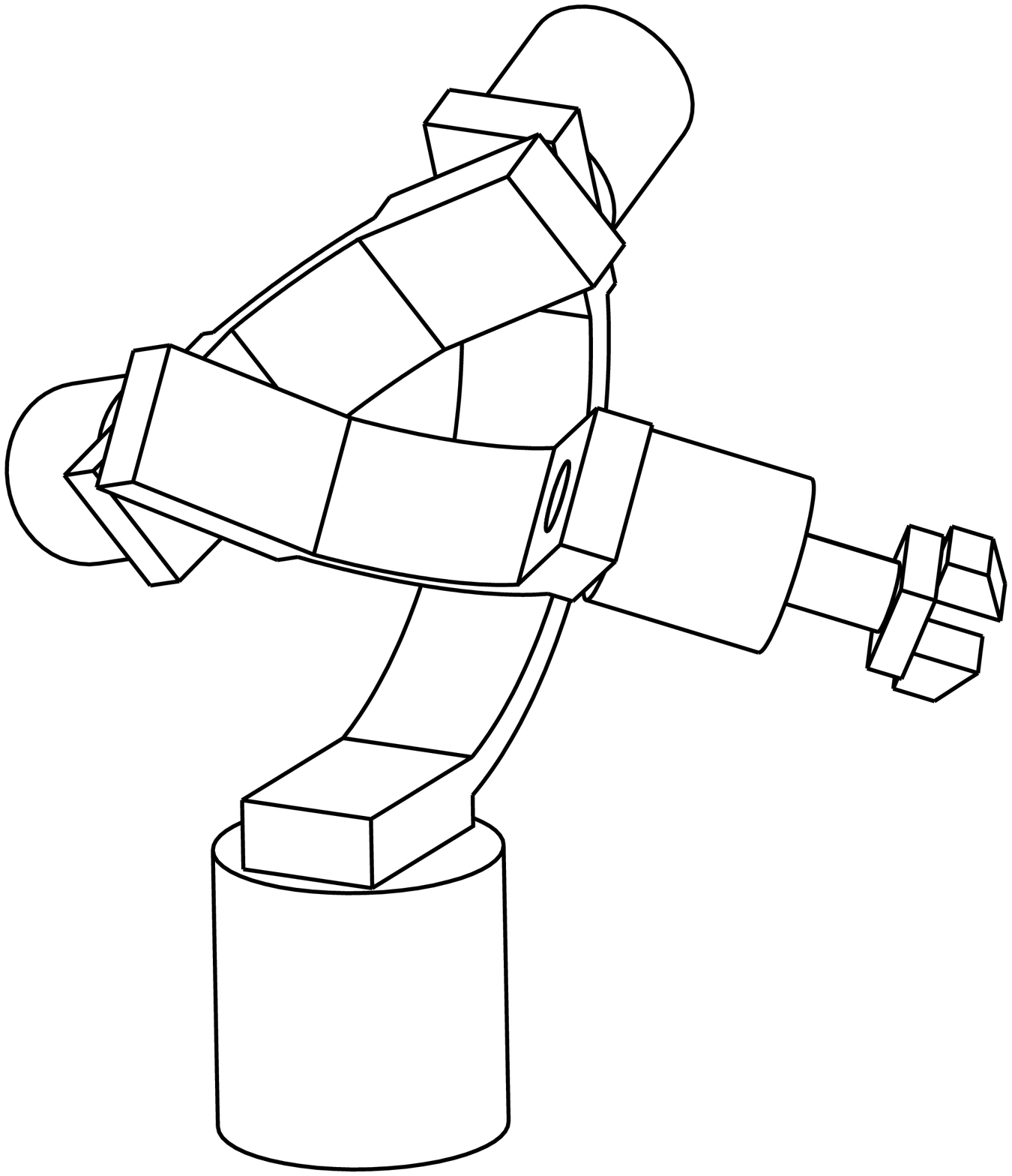}}}
           \centerline{\hbox{(c)}}
       \end{minipage} &
       \begin{minipage}[t]{30 mm}
           \centerline{\hbox{\includegraphics[width=30mm,height=45mm]{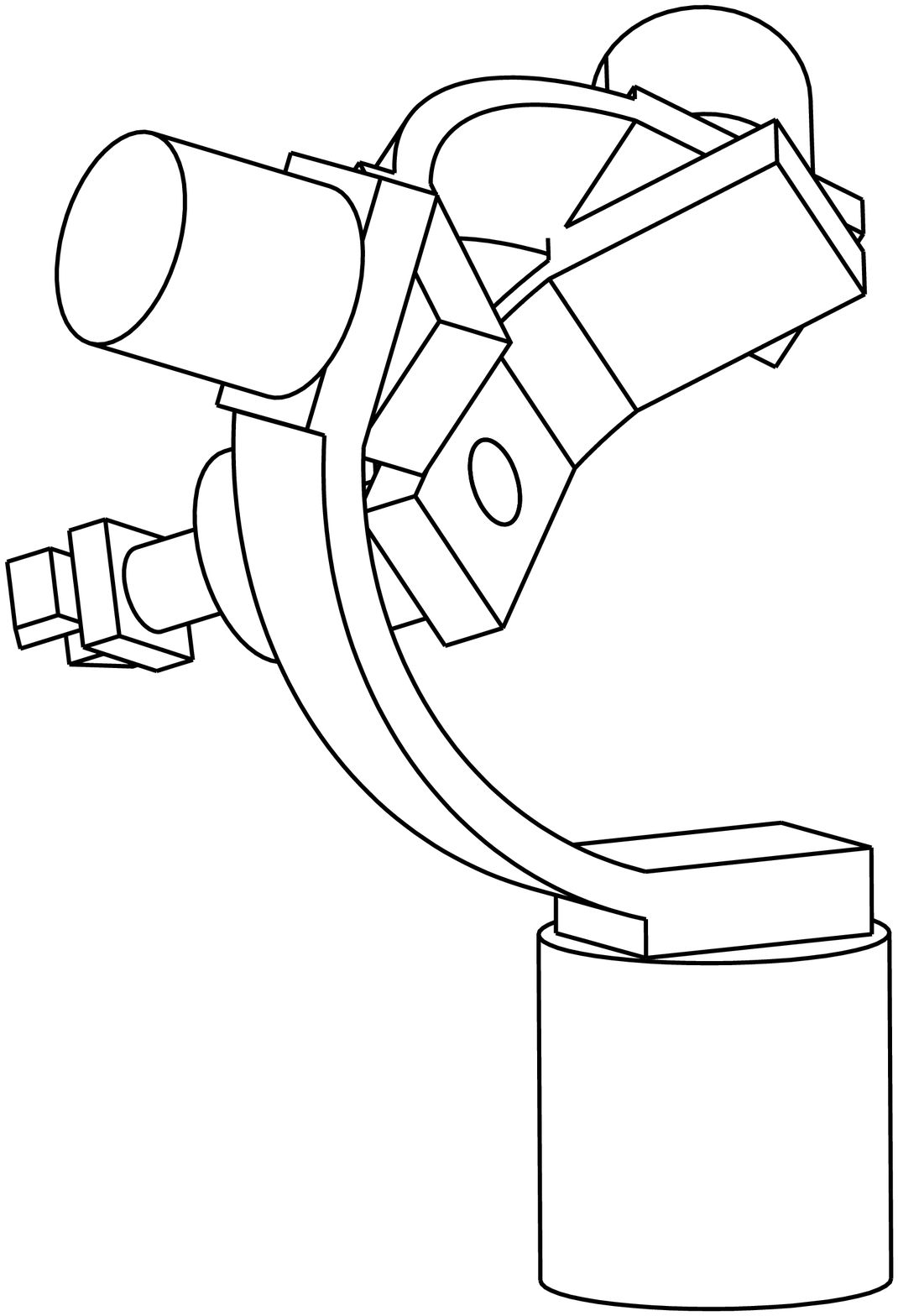}}}
           \centerline{\hbox{(d)}}
       \end{minipage} \\
       \begin{minipage}[t]{33 mm}
           \centerline{\hbox{\includegraphics[width=37mm,height=41mm]{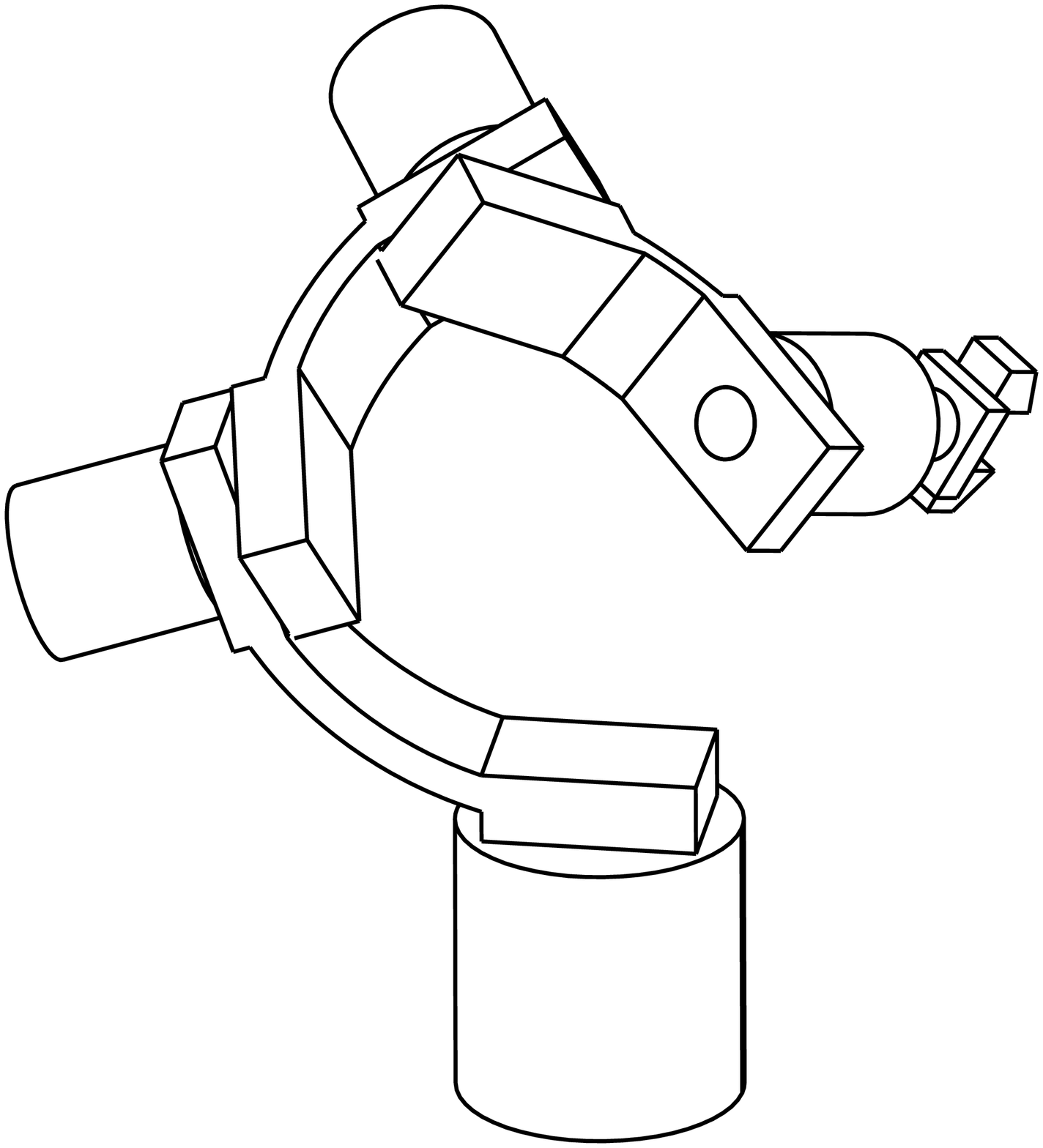}}}
           \centerline{\hbox{(e)}}
       \end{minipage} &
       \begin{minipage}[t]{33 mm}
           \centerline{\hbox{\includegraphics[width=35mm,height=38mm]{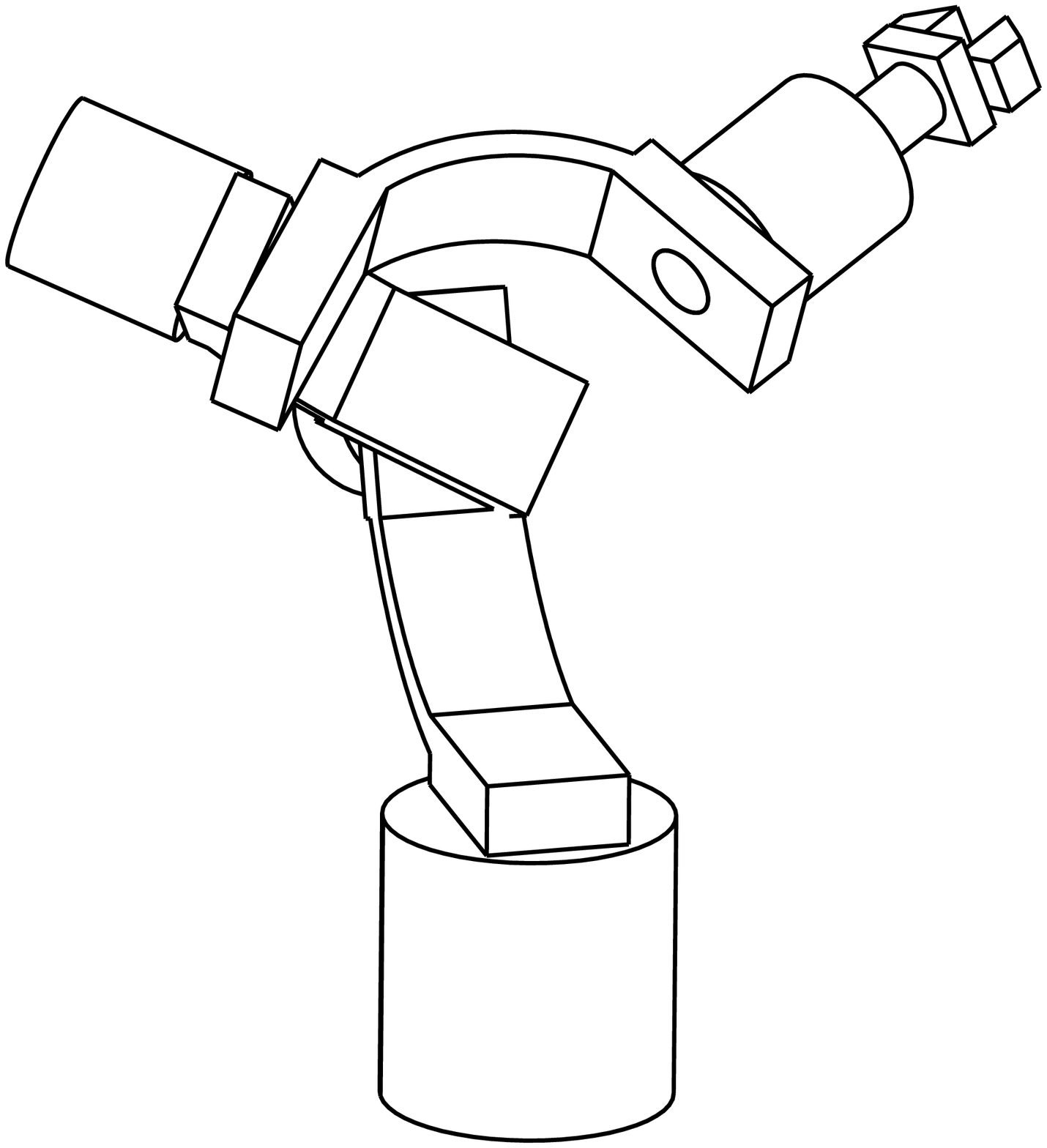}}}
           \centerline{\hbox{(f)}}
       \end{minipage} &
       \begin{minipage}[t]{33 mm}
           \centerline{\hbox{\includegraphics[width=37mm,height=44mm]{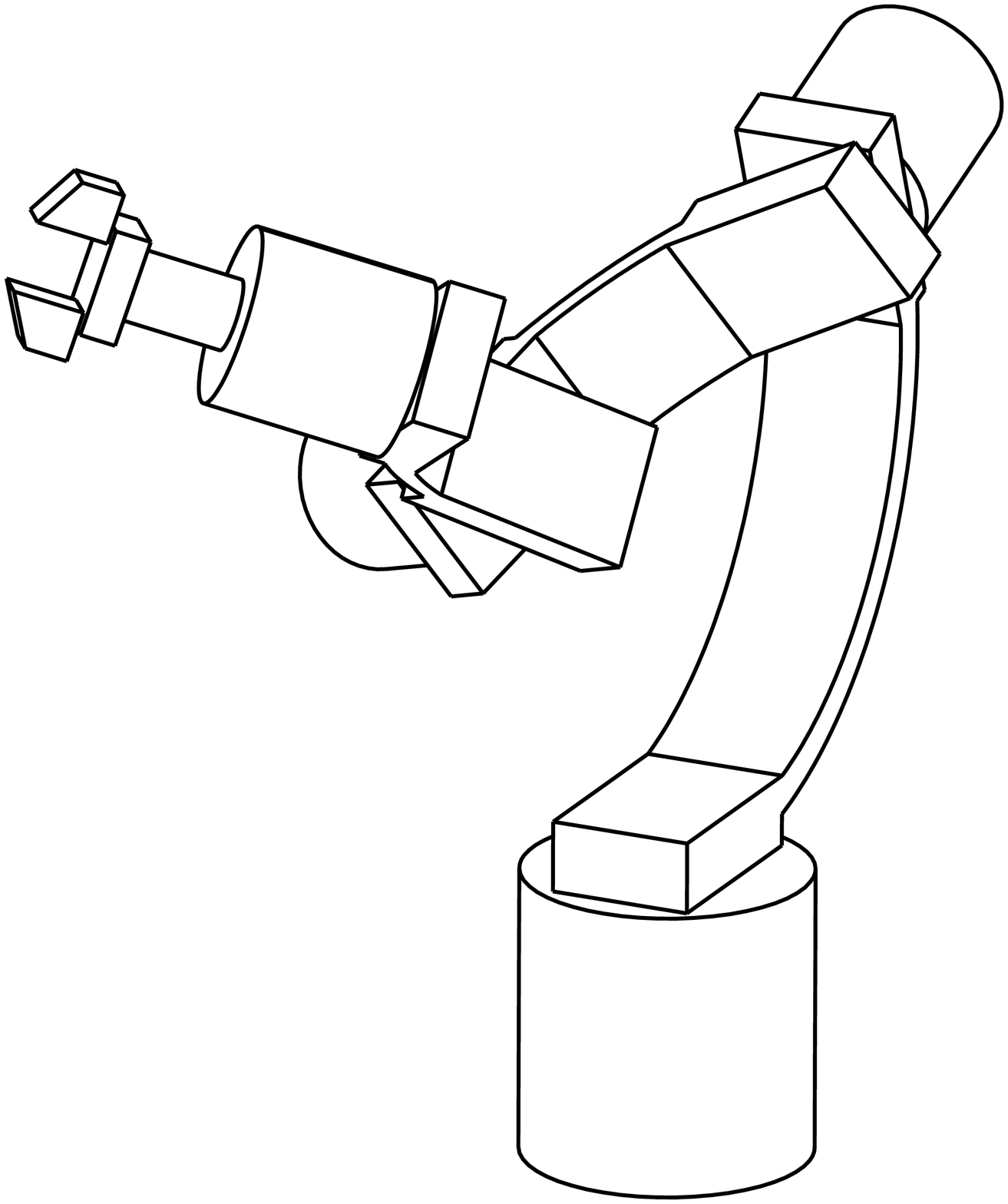}}}
           \centerline{\hbox{(g)}}
       \end{minipage} &
       \begin{minipage}[t]{30 mm}
           \centerline{\hbox{\includegraphics[width=30mm,height=46mm]{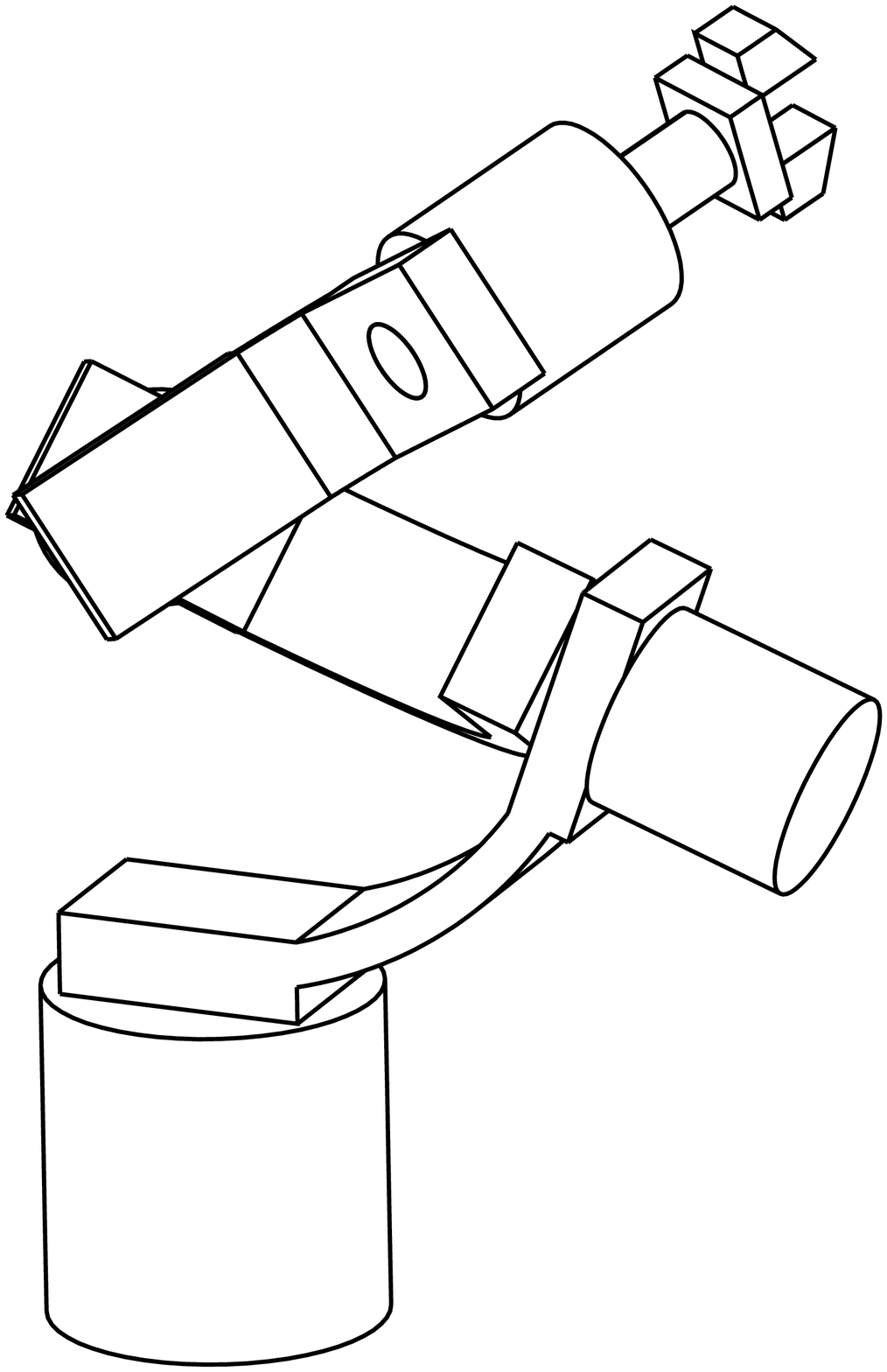}}}
           \centerline{\hbox{(h)}}
       \end{minipage}
    \end{tabular}
    \end{center}
    \caption{The eight distincts isotropic wrists of Table~5}
    \protect\label{figure:Poignet_C1}
\end{figure}

Now, for all seven solutions of Table~\ref{Table:Antipodal} and
the trivial isotropic set of points given in
eq.(\ref{equation:trivial_set}), we compute the corresponding
Denavit-Hartenberg (DH) parameters yielding isotropic wrists. For
each isotropic set of points, we place the first joint axis at
$P_1$. It is now apparent that we can derive six kinematic chains.
Thus, we find $\alpha_1$, $\alpha_2$ and $\alpha_3$ as the angles
made by the neighboring position vectors of points $P_i$.
Moreover, we eliminate the set of DH parameters leading to wrists
that are identical. Hence, a total of eight distinct isotropic
wrists are obtained from these sets. The Denavit-Hartenberg
parameters of the eight distinct wrists are displayed in
Table~\ref{Table:DH_Parameters}. The corresponding wrists, at
their isotropic postures, being displayed in
Figs.~\ref{figure:Poignet_C1}a--h.

\begin{table}[!hb]
  \begin{center}
     \begin{tabular}{|c|c|c|} \hline
       $i$ & $\alpha_i$     & $\theta_i$        \\ \hline
        1  & $109.5^\circ$  & $\theta_1$        \\ \hline
        2  & $109.5^\circ$  & $\pm 60^\circ$    \\ \hline
        3  & $109.5^\circ$  & $\pm (-60^\circ)$ \\ \hline
        4  & $\star$        & $\theta_4$        \\ \hline
        \multicolumn{3}{c}{(a)}
     \end{tabular}
     \begin{tabular}{|c|c|c|} \hline
       $i$ & $\alpha_i$     & $\theta_i$      \\ \hline
        1  & $70.5^\circ$   & $\theta_1$      \\ \hline
        2  & $109.5^\circ$  & $\pm 120^\circ$ \\ \hline
        3  & $109.5^\circ$  & $\pm 60^\circ$  \\ \hline
        4  & $\star$        & $\theta_4$      \\ \hline
        \multicolumn{3}{c}{(b)}
     \end{tabular}
     \begin{tabular}{|c|c|c|} \hline
       $i$ & $\alpha_i$     & $\theta_i$      \\ \hline
        1  & $109.5^\circ$  & $\theta_1$      \\ \hline
        2  & $70.5^\circ$   & $\pm 120^\circ$ \\ \hline
        3  & $109.5^\circ$  & $\pm 120^\circ$ \\ \hline
        4  & $\star$        & $\theta_4$      \\ \hline
        \multicolumn{3}{c}{(c)}
     \end{tabular}
     \begin{tabular}{|c|c|c|} \hline
       $i$ & $\alpha_i$     & $\theta_i$      \\ \hline
        1  & $109.5^\circ$  & $\theta_1$      \\ \hline
        2  & $109.5^\circ$  & $\pm 60^\circ$  \\ \hline
        3  & $70.5^\circ$   & $\pm 120^\circ$ \\ \hline
        4  & $\star$        & $\theta_4$      \\ \hline
        \multicolumn{3}{c}{(d)}
     \end{tabular}
     \begin{tabular}{|c|c|c|} \hline
       $i$ & $\alpha_i$     & $\theta_i$     \\ \hline
        1  & $70.5^\circ$   & $\theta_1$     \\ \hline
        2  & $70.5^\circ$   & $\pm 60^\circ$ \\ \hline
        3  & $70.5^\circ$   & $\pm 60^\circ$ \\ \hline
        4  & $\star$        & $\theta_4$     \\ \hline
        \multicolumn{3}{c}{(e)}
     \end{tabular}
     \begin{tabular}{|c|c|c|} \hline
       $i$ & $\alpha_i$     & $\theta_i$              \\ \hline
        1  & $70.5^\circ$   & $\theta_1$              \\ \hline
        2  & $70.5^\circ$   & $\pm 60^\circ$          \\ \hline
        3  & $109.5^\circ$  & $\!\pm (-120^\circ)\!$  \\ \hline
        4  & $\star$        & $\theta_4$              \\ \hline
        \multicolumn{3}{c}{(f)}
     \end{tabular}
     \begin{tabular}{|c|c|c|} \hline
       $i$ & $\alpha_i$     & $\theta_i$              \\ \hline
        1  & $109.5^\circ$  & $\theta_1$              \\ \hline
        2  & $70.5^\circ$   & $\pm 120^\circ$         \\ \hline
        3  & $70.5^\circ$   & $\!\pm (-60^\circ)\!\!$ \\ \hline
        4  & $\star$        & $\theta_4$              \\ \hline
        \multicolumn{3}{c}{(g)}
     \end{tabular}
     \begin{tabular}{|c|c|c|} \hline
       $i$ & $\alpha_i$     & $\theta_i$               \\ \hline
        1  & $70.5^\circ$   & $\theta_1$               \\ \hline
        2  & $109.5^\circ$  & $\pm 120^\circ$          \\ \hline
        3  & $70.5^\circ$   & $\!\pm (-120^\circ)\!\!$ \\ \hline
        4  & $\star$        & $\theta_4$               \\ \hline
        \multicolumn{3}{c}{(h)}
     \end{tabular}
  \caption{The Denavit-Hartenberg parameters of the eight spherical
wrists at their isotropic postures}
  \label{Table:DH_Parameters}
  \end{center}
\end{table}

Note that the entry corresponding to $\alpha_4$ in the foregoing
table is left with an asterisk because this twist angle is not
defined for a four-revolute wrist. Its value depends on how the
$z$-axis of the task frame is defined. As well, angles $\theta_1$
and $\theta_4$ are left unspecified because isotropy is
independent of these values, i.e., isotropy is preserved upon
varying these two angles throughout their whole range of values,
from 0 to $2\pi$.
%%%%%%%%%%%%%%%%%%%%%%%%%%%%%%
\section{Conclusions}
%%%%%%%%%%%%%%%%%%%%%%%%%%%%%%
We showed that the algebraic formulation of the problem leading to
all four-revolute serial spherical wrists with kinetostatic
isotropy yields a system of eight quadratic equations in eight
unknowns, whose Bezout number is 256, its BKK bound being $192$.
Nevertheless, this system admits only 32 distinct solutions.
Furthermore, upon elimination of the solutions leading to repeated
wrists, we are left with only eight distinct isotropic wrists,
whose Denavit-Hartenberg parameters were computed and displayed,
the corresponding wrists having been displayed at their isotropic
postures.
%%%%%%%%%%%%%%%%%%%%%%%%%%%%%%
\bibliographystyle{unsrt}
%%%%%%%%%%%%%%%%%%%%%%%%%%%%%%

%%%%%%%%%%%%%%%%%%%%%%%%%%%
\section*{Appendix}
% \begin{figure}[hbt]
\begin{floatingfigure}[r]{45mm}
  \begin{center}
    \centerline{\hbox{\includegraphics[width=36mm,height=31mm]{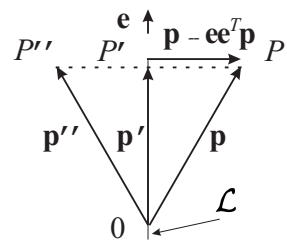}}}
   \caption{The reflection of a point with respect to a line}
   \protect\label{f:ref_line}
  \end{center}
\end{floatingfigure}
% \end{figure}
We show here that a reflection \negr L about a line $\cal L$ that
passes through the origin is a rotation about $\cal L$ through an
angle $\pi$. To this end, we resort to Fig.~\ref{f:ref_line},
showing $\cal L$ and a point $P$. To simplify matters, we sketch
$P$ and $\cal L$ in their plane.

The {\em projection} of $P$ onto $\cal L$ is denoted by
$P^\prime$, the reflection sought by $P^{\prime\prime}$, the
corresponding position vectors being denoted by \negr p, $\negr
p^\prime$ and $\negr p^{\prime\prime}$. Apparently,
 \beq \negr
   p^\prime = \negr e\negr e^T\negr p\label{e:p'}
 \eeq

Hence,
 \bed
  \negr p^{\prime\prime}=\negr p^\prime - (\negr p -
  \negr e\negr e^T\negr p)=(2\negr e\negr e^T -\negr 1)\negr p
 \eed
Thus, the reflection sought, \negr L, is given by the matrix
coefficient of \negr p in the rightmost side of the foregoing
equation, i.e., \beq \negr L = 2\negr e\negr e^T -\negr
1\label{e:L} \eeq As the reader can readily verify, the above
expression yields a proper orthogonal matrix, and hence, a
rotation. Moreover, the axis of the rotation is \negr e and the
angle $\pi$.
\end{document}